\documentclass{article}
\usepackage{mdframed}
\usepackage{microtype}
\usepackage{amsmath}
\usepackage{hyperref}
\usepackage{amsfonts}
\usepackage{xcolor}
\usepackage{adjustbox}
\usepackage{booktabs}
\usepackage{bbm}
\usepackage{caption}
\usepackage{subcaption}

\usepackage[final]{neurips_2020}

\newcommand{\inv}{\text{inv}}
\newcommand{\spu}{\text{spu}}
\newcommand{\env}{\text{env}}

\usepackage{enumitem}
\setlist{leftmargin=5mm}

\hypersetup{
    colorlinks,
    linkcolor={blue!50!black},
    citecolor={blue!50!black},
    urlcolor={blue!50!black}
}

\title{Linear unit-tests for invariance discovery}

\author{
Benjamin Aubin \\
Facebook AI Research\\
Paris, France\\
\And
Agnieszka S\l{}owik \\
Facebook AI Research\\
London, UK
\And
Martin Arjovsky \\
INRIA - PSL Research University\\
Paris, France
\And
Leon Bottou \\
Facebook AI Research,\\
New York, NY, 10003, USA
\And
David Lopez-Paz \\
Facebook AI Research\\
Paris, France\\
\texttt{dlp@fb.com}
}

\begin{document}

\maketitle

\begin{abstract}
There is an increasing interest in algorithms to learn invariant correlations across training environments.
A big share of the current proposals find theoretical support in the causality literature but, how useful are they in practice? 
The purpose of this note is to propose six linear low-dimensional problems ---``unit tests''--- to evaluate different types of out-of-distribution generalization in a precise manner. 
Following initial experiments, none of the three recently proposed alternatives passes all tests. 
By providing the code to automatically replicate all the results in this manuscript ({\url{https://www.github.com/facebookresearch/InvarianceUnitTests}}), we hope that our unit tests become a standard stepping stone for researchers in out-of-distribution generalization.
\end{abstract}

\section{Introduction}

Machine learning systems crumble when deployed in conditions different to those of training \citep{szegedy2013intriguing, rosenfeld2018elephant, alcorn2019strike}. 
To address this issue, recent works in causality \citep{peters2015causal, arjovsky2019invariant, parascandolo2020learning} propose to learn correlations invariant across multiple training distributions, and to use those correlations as a proxy for out-of-distribution generalization.
However, as we will show, these algorithms perform poorly across a catalog of simple low-dimensional linear problems.
Therefore, our contribution is a standardized set of six ``unit tests'' that researchers can bear in mind when proposing new solutions for out-of-distribution generalization.

Causal learning algorithms such as Invariant Causal Prediction \citep[ICP]{peters2015causal} and Invariant Risk Minimization \citep[IRM]{arjovsky2019invariant} consume
several training datasets, each of them possibly produced by the same ``structural equation model'' operating under a different ``valid interventions''.
A structural equation model is a list of equations describing how variables influence each other to take their values \citep{pearl2009causality, peters2017elements}.
An intervention perturbs one or more of these equations, and it is ``valid'' as long as it does not modify the conditional expectation of the target variable given its direct causal parents \citep{arjovsky2019invariant}.
Then, if the interventions producing our training datasets are diverse, invariant correlations should pertain to the fixed causal mechanism of the target variable.
This suggests the possibility of learning about the causal structure of data by searching for statistical invariances.
In turn, these invariances enable out-of-distribution generalization: 
if the causal mechanism of the target variable is invariant across diverse training conditions, we may rely on them to perform robustly at novel test conditions.

Unfortunately, invariances are often more difficult to capture than spurious correlations.
As an example, consider the task of classifying pictures of cows and camels \citep{arjovsky2019invariant}.
A ``green detector'' solves this task to a great extent, since almost all pictures of cows contain green pastures, and almost all pictures of camels show beige sandy landscapes.
Following the principle of least effort \citep{geirhos2020shortcut}, learning machines cling to the textural ``green-cow'' spurious correlation, rather than discovering the shapes that make a cow a cow.
Predictors absorbing these ``distractor'', ``shortcut'', or ``bait'' correlations fail when deployed under novel conditions.

Although designed to address this very issue, causal learning algorithms often fail to capture causal invariances in data. 
The purpose of this note is to share six linear problems that illustrate this phenomena.
Each of these problems contains an invariant causal correlation (\emph{inv}) that we would like to learn, as well as a spurious correlation (\emph{spu}) that we would like to discard.
In every dataset, empirical risk minimization absorbs the spurious correlation from the training data, failing to generalize to novel conditions.
By releasing the code to automatically replicate all the results in this manuscript,
we hope that these ``unit tests'' become a standard guide when developing new out-of-distribution generalization algorithms.

\section{Problems}
\label{problems}

For each problem, we collect datasets $D_{e} = \{(x_i^e, y_i^e)\}_{i=1}^{n_{e}}$ containing $n_{e}$ samples for $n_\env$ environments $e \in \mathcal{E} = \{E_j\}_{j=1}^{n_\env}$.
The input feature vector $x^e = (x^e_{\text{inv}}, x^e_{\text{spu}}) \in \mathbb{R}^{d}$ contains features $x^e_{\text{inv}} \in \mathbb{R}^{d_\inv} $ that elicit invariant correlations, as well as features $x^e_{\text{spu}} \in  \mathbb{R}^{d_\spu}$ that elicit spurious correlations.
Our goal is to construct invariant predictors that estimate the target variable $y^e$ by relying on $x^e_\text{inv}$, while ignoring $x^e_{\text{spu}}$.
To measure the extent to which an algorithm ignores the features $x^e_{\text{spu}}$, we sample a \emph{train} split, a \emph{validation} split, and a \emph{test} split per problem and environment.
Both train and validation splits are built by sampling the structural equations outlined below for each problem and environment.
The test split is built analogously, but the features $x^e_{\text{spu}}$ are shuffled at random across examples.
This way, only those predictors ignoring the shortcut correlations provided by $x^e_{\text{spu}}$ will achieve minimal test error.

Finally, let $\mathcal{N}_d(\mu, \sigma^2)$ be the $d$-dimensional Gaussian distribution with mean vector $\mu$ and diagonal covariance matrix with diagonal elements $\sigma^2$. We fix $n_{e} = 10^4$ and we mainly illustrate our results in the default case $n_\env=3$ with $(d, d_\inv, d_\spu)=(10, 5, 5)$.

\subsection{\texttt{Example1}: regression from causes and effects}%
\label{sub:problem_1_regression_from_causes_and_effects}

A linear least-squares regression problem where features contain causes and effects of the target variable, as proposed by \citet{arjovsky2019invariant}.
By virtue of considering only valid interventions, the mapping from the causes to the target variable is invariant across environments.
Conversely, the mapping from the target variable to the effects changes across environments. 
To construct the datasets $D_{e}$ for every $e \in \mathcal{E}$ and $i = 1, \ldots, n_{e}$, sample:
\begin{center}
\vspace{-0.5cm}
\begin{minipage}[c]{0.35\linewidth}
\begin{align*}
    x^e_{\text{inv}, i} &\sim \mathcal{N}_{d_\inv}(0, (\sigma^e)^2),\\
    \tilde{y}^e_i &\sim \mathcal{N}_{d_\inv}( W_{yx} ~ x^e_{\text{inv}, i}, (\sigma^e)^2),\\
    x^e_{\text{spu}, i} &\sim \mathcal{N}_{d_\spu}( W_{xy} ~ \tilde{y}^e_i, 1),
\end{align*}
\end{minipage}
\begin{minipage}[c]{0.35\linewidth}
\begin{align*}
    x^e_i &\leftarrow (x_{\text{inv}, i}^e, x_{\text{spu}, i}^e),\\
    y^e_i &\leftarrow \frac{2}{d} \cdot 1_{d_\inv}^\top ~ \tilde{y}^e_i;
\end{align*}
\end{minipage}
\end{center}
where the matrices $W_{yx} \in \mathbb{R}^{d_\inv \times d_\inv}$, $W_{xy} \in \mathbb{R}^{d_\spu \times d_\inv}$ are drawn i.i.d from the Gaussian normal distribution. 
The first environment variables are fixed to $(\sigma^{e=E_0})^2 = 0.1$, $(\sigma^{e=E_1})^2 = 1.5$, and $(\sigma^{e=E_2})^2 = 2$. For $n_\env >3$ and $j \in [3 : n_\env-1]$, the extra environments $(\sigma^{e=E_j})^2$ are drawn uniformly from $\textrm{Unif}(10^{-2}, 10)$.

\paragraph{Challenges} 
First, the distribution of the feature $x^e_\text{inv}$ eliciting the invariant correlation changes across environments.
This disallows the use of domain-adversarial methods \citep{ganin2016domain}, which seek features with matching distribution across environments.
Second, the distribution of the residuals varies across environments, disallowing the use of ICP \citep{peters2015causal}.
%
Third, since the target variable is continuous, conditional domain-adversarial techniques \citep{li2018deep} do not apply easily.
This example is solved by IRM, and it is analogous to a linear regression variant of the ColorMNIST task \citep{arjovsky2019invariant}.

\subsection{\texttt{Example2}: cows versus camels}%
\label{sub:problem_2_cows_versus_camels}

In the spirit of \citep{beery2018recognition, arjovsky2019invariant}, we add a
binary classification problem to imitate the introductory example ``most cows appear in grass and most camels appear in sand''. Let:
\begin{align*}
    \mu_{\text{cow}} &\sim 1_{d_\inv},&&  \mu_{\text{camel}} = - \mu_{\text{cow}}, && \nu_\text{animal} = 10^{-2}\,, \\
    \mu_{\text{grass}} &\sim 1_{d_\spu},&&  \mu_{\text{sand}} = - \mu_{\text{grass}}, && \nu_\text{background} = 1.
\end{align*}

To construct the datasets $D_{e}$ for every $e \in \mathcal{E}$ and $i = 1, \ldots, n_{e}$, sample:
\begin{align*}
    j^e_i &\sim \text{Categorical}\left(p^e s^e , (1 - p^e) s^e, p^e (1-s^e) , (1 - p^e) (1-s^e)\right);
\end{align*}
\begin{minipage}[c]{0.61\linewidth}
    \begin{align*}
        x^e_{\text{inv}, i} &\sim
        \left\{\begin{array}{lr}
            (\mathcal{N}_{d_\inv}(0, 10^{-1}) + \mu_\text{cow}) \cdot \nu_\text{animal} & \text{ if } j^e_i \in \{1, 2\},\\
            (\mathcal{N}_{d_\inv}(0, 10^{-1}) + \mu_\text{camel}) \cdot \nu_\text{animal} & \text{ if } j^e_i \in \{3, 4\},\\
        \end{array}\right.\\
        x^e_{\text{spu}, i} &\sim
        \left\{\begin{array}{lr}
            (\mathcal{N}_{d_\spu}(0, 10^{-1}) + \mu_\text{grass}) \cdot \nu_\text{background} & \text{ if } j^e_i \in \{1, 4\},\\
            (\mathcal{N}_{d_\spu}(0, 10^{-1}) + \mu_\text{sand})  \cdot \nu_\text{background} & \text{ if } j^e_i \in \{2, 3\},
        \end{array}\right. 
    \end{align*}
\end{minipage}
\hfill
\begin{minipage}[c]{0.32\linewidth}
    \begin{align*}
        x^e_i &\leftarrow (x^e_{\text{inv}, i}, x^e_{\text{spu}, i});\\
        y^e_i &\leftarrow
        \left\{\begin{array}{lr}
        1 & \text{ if } 1_{d_\inv}^\top x^e_{i, \text{inv}} > 0,\\
        0 & \text{else};
        \end{array}\right.
        \end{align*}
\end{minipage}

where the environment foreground/background probabilities are $p^{e=E_0} = 0.95$, $p^{e=E_1} = 0.97$, $p^{e=E_2} = 0.99$ and the cow/camel probabilities are $s^{e=E_0} = 0.3$, $s^{e=E_1} = 0.5$, $s^{e=E_2} = 0.7$. For $n_\env >3$ and $j \in [3 : n_\env-1]$, the extra environment variables are respectively drawn according to $p^{e=E_j} \sim \textrm{Unif}(0.9, 1)$ and $s^{e=E_j} \sim \textrm{Unif}(0.3, 0.7)$.

\paragraph{Challenges}
Achieving zero \emph{population} error while using only $x^e_\text{inv}$ requires learning large weights.
This is difficult when using gradient descent methods, or most forms of regularization.
As the dimension of the feature space grows, the probability of achieving zero \emph{training} error using only $x^e_\text{spu}$ increases rapidly.
This means that invariance penalties based on training error, such as IRM, may accept solutions using spurious features.

\subsection{\texttt{Example3}: small invariant margin}%
\label{sub:problem_3_bait_margin}

A linear version of the spiral binary classification problem proposed by \citep{parascandolo2020learning}, where
the first two dimensions offer an invariant, small-margin linear decision boundary.
The rest of the dimensions offer a changing, large-margin linear decision boundary.
Let $\gamma = 0.1 \cdot 1_{d_\inv}$, and $\mu^e \sim \mathcal{N}_{d_\spu}(0, 1)$, for all environments.

To construct the datasets $D_{e}$ for every $e \in \mathcal{E}$ and $i = 1, \ldots, n_{e}$, sample:\\
\begin{minipage}[c]{0.69\linewidth}
    \begin{align*}
        y^e_i &\sim \text{Bernoulli}\left(\frac{1}{2}\right),\\
        x^e_{\text{inv}, i} &\sim
        \left\{\begin{array}{lr}
            \mathcal{N}_{d_\inv}(+\gamma, 10^{-1}) & \text{ if } y^e_i = 0,\\
            \mathcal{N}_{d_\inv}(-\gamma, 10^{-1}) & \text{ if } y^e_i = 1;\\
        \end{array}\right.\\
        x^e_{\text{spu}, i} &\sim
        \left\{\begin{array}{lr}
            \mathcal{N}_{d_\spu}(+\mu^e, 10^{-1}) & \text{ if } y^e_i = 0,\\
            \mathcal{N}_{d_\spu}(-\mu^e, 10^{-1}) & \text{ if } y^e_i = 1;\\
        \end{array}\right.
    \end{align*}
\end{minipage}
\begin{minipage}[c]{0.05\linewidth}
    \begin{align*}
        x^e_i &\leftarrow (x^e_{\text{inv}, i}, x^e_{\text{spu}, i}).
        \end{align*}
\end{minipage}

\paragraph{Challenges} 
We can solve this problem to zero \emph{population} error with high probability using $x^e_\text{spu}$ alone.
Things complicate further, as solving this task using $x^e_\text{inv}$ forcefully incurs a small amount of \emph{population} error. 
Therefore, learning algorithms should learn to sacrifice training error to realize that $x^e_\text{inv}$ lead to the same \emph{maximum margin} classifier across environments (even though the varying margin based on $x^e_\text{spu}$ is larger!).
While the predictor based on $x^e_\text{inv}$ is the optimal in terms of worst-case out-of-distribution generalization, it is not a causal predictor of the target variable.

\subsection{Scrambled variations}%
\label{sub:scrambled_problem_variations}

We define three additional problems: \texttt{example1s}, \texttt{example2s}, and \texttt{example3s}.
These are ``scrambled'' variations of the three problems described above, respectively.
Scrambled variations build observed datasets $D^e = \{(S^\top x^e_i, y^e_i)\}_{i=1}^{n^e}$, where $S \in \mathbb{R}^{d\times d}$ is a random rotation matrix fixed for all environments $e\in\mathcal{E}$.
Observing a scrambled version of the variables appearing in the structural equation models requires algorithms to learn a (linear) feature representation under which the desired invariance should be elicited.
Because of this reason, we do not compare to brute-force feature selection methods, such as ICP \citep{peters2015causal}.


\section{Experiments}

We provide an initial set of experiments evaluating the following algorithms on our six problems:
Empirical Risk Minimization \citep[ERM]{vapnik1998statistical} minimizes the error on the union of all the training splits. 
Invariant Risk Minimization \citep[IRMv1]{arjovsky2019invariant} finds a representation of the features such that the optimal classifier, on top of that representation, is the identity function for all environments.
Inter-environmental Gradient Alignment \citep[IGA]{koyama2020out} minimizes the error on the training splits while reducing the variance of the gradient of the loss per environment.
AND-mask \citep{parascandolo2020learning} minimizes the error on the training splits by updating the model on those directions where the sign of the gradient of the loss is the same for most environments. 
Oracle is a version of ERM where all data splits contain randomized $x^e_\text{spu}$, and therefore are trivial to ignore. The purpose of this method is to understand the achievable upper bound performance in our problems. 

For each algorithm, we run a random hyper-parameter search of 20 trials.
We trained each algorithm and hyper-parameter trial on the train splits of all environments, for $10^4$ full-batch Adam \citep{kingma2014adam} updates.
We choose the hyper-parameters trial that minimizes the error on the validation splits of all environments. 
Finally, we report the error of these selected models on the test splits.
To provide error bars, this entire process, including data sampling, is repeated 50 times.
We refer the reader to our code to learn about the hyper-parameter search distributions for each algorithm.

\begin{table}
\caption{Test errors for all algorithms, datasets, and environments for $(d_\inv, d_\spu, n_\env)=(5,5,3)$.
Example 1 and Example 1s errors are in MSE; all others are classification errors.}
\vskip 0.2cm
\centering
\adjustbox{max width=0.8\textwidth}{%
\begin{tabular}{lccccc}
    \toprule
                    & ANDMask         & ERM             & IGA             & IRMv1           & Oracle          \\
    \midrule
    Example1.E0     & 0.11 $\pm$ 0.04 & 1.62 $\pm$ 0.60 & 4.47 $\pm$ 1.16 & 0.20 $\pm$ 0.04 & 0.05 $\pm$ 0.00 \\
    Example1.E1     & 11.39 $\pm$ 0.18 & 14.25 $\pm$ 1.52 & 18.46 $\pm$ 2.14 & 11.98 $\pm$ 0.75 & 11.27 $\pm$ 0.17 \\
    Example1.E2     & 20.28 $\pm$ 0.30 & 24.22 $\pm$ 2.34 & 29.48 $\pm$ 3.19 & 21.27 $\pm$ 1.34 & 19.93 $\pm$ 0.31 \\
    Example1s.E0    & 0.07 $\pm$ 0.01 & 1.61 $\pm$ 0.59 & 4.55 $\pm$ 1.79 & 0.19 $\pm$ 0.04 & 0.05 $\pm$ 0.00 \\
    Example1s.E1    & 12.13 $\pm$ 0.80 & 14.23 $\pm$ 1.49 & 18.68 $\pm$ 3.37 & 11.92 $\pm$ 0.69 & 11.24 $\pm$ 0.19 \\
    Example1s.E2    & 21.52 $\pm$ 1.42 & 24.14 $\pm$ 2.39 & 29.81 $\pm$ 4.78 & 21.08 $\pm$ 1.31 & 20.06 $\pm$ 0.37 \\
    \midrule
    Example2.E0     & 0.42 $\pm$ 0.02 & 0.40 $\pm$ 0.01 & 0.43 $\pm$ 0.00 & 0.43 $\pm$ 0.00 & 0.00 $\pm$ 0.00 \\
    Example2.E1     & 0.49 $\pm$ 0.03 & 0.47 $\pm$ 0.01 & 0.50 $\pm$ 0.00 & 0.50 $\pm$ 0.00 & 0.00 $\pm$ 0.00 \\
    Example2.E2     & 0.42 $\pm$ 0.02 & 0.40 $\pm$ 0.01 & 0.42 $\pm$ 0.01 & 0.42 $\pm$ 0.01 & 0.00 $\pm$ 0.00 \\
    Example2s.E0    & 0.43 $\pm$ 0.01 & 0.43 $\pm$ 0.01 & 0.43 $\pm$ 0.01 & 0.43 $\pm$ 0.01 & 0.00 $\pm$ 0.00 \\
    Example2s.E1    & 0.50 $\pm$ 0.00 & 0.50 $\pm$ 0.00 & 0.50 $\pm$ 0.00 & 0.50 $\pm$ 0.00 & 0.00 $\pm$ 0.00 \\
    Example2s.E2    & 0.42 $\pm$ 0.01 & 0.42 $\pm$ 0.01 & 0.42 $\pm$ 0.01 & 0.42 $\pm$ 0.01 & 0.00 $\pm$ 0.00 \\
    \midrule
    Example3.E0     & 0.35 $\pm$ 0.22 & 0.48 $\pm$ 0.09 & 0.47 $\pm$ 0.10 & 0.49 $\pm$ 0.07 & 0.00 $\pm$ 0.00 \\
    Example3.E1     & 0.36 $\pm$ 0.22 & 0.48 $\pm$ 0.07 & 0.48 $\pm$ 0.08 & 0.49 $\pm$ 0.06 & 0.00 $\pm$ 0.00 \\
    Example3.E2     & 0.32 $\pm$ 0.22 & 0.47 $\pm$ 0.12 & 0.46 $\pm$ 0.12 & 0.48 $\pm$ 0.07 & 0.00 $\pm$ 0.00 \\
    Example3s.E0    & 0.45 $\pm$ 0.13 & 0.48 $\pm$ 0.08 & 0.48 $\pm$ 0.09 & 0.49 $\pm$ 0.07 & 0.00 $\pm$ 0.00 \\
    Example3s.E1    & 0.49 $\pm$ 0.05 & 0.49 $\pm$ 0.05 & 0.48 $\pm$ 0.07 & 0.49 $\pm$ 0.06 & 0.00 $\pm$ 0.00 \\
    Example3s.E2    & 0.46 $\pm$ 0.12 & 0.47 $\pm$ 0.09 & 0.47 $\pm$ 0.11 & 0.48 $\pm$ 0.07 & 0.00 $\pm$ 0.00 \\
    \bottomrule
    \end{tabular}
}
\label{fig:averaged_results}
\label{main_results}
\end{table}

\begin{figure}
    \centering
    \includegraphics[width=\linewidth]{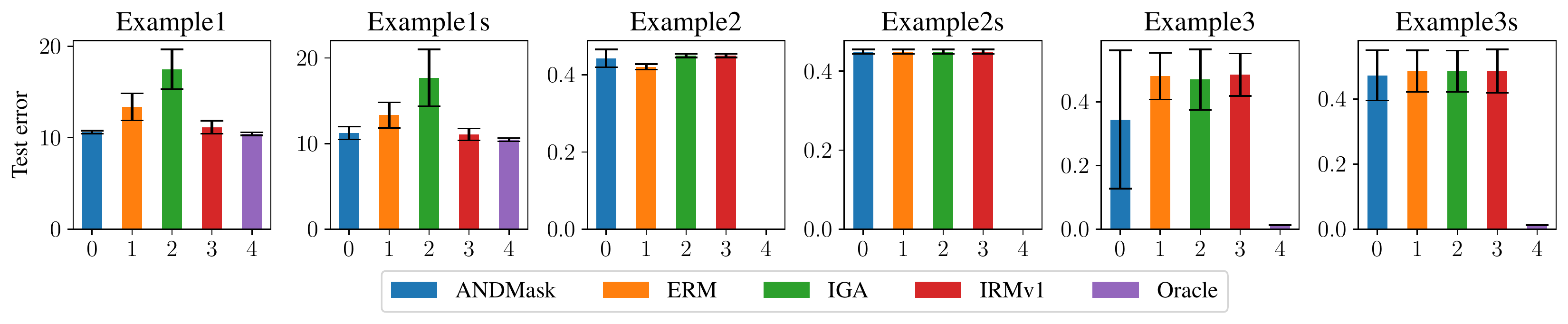}
    \captionof{figure}{Test error averaged across environments (E0, E1, E2) for $(d_\inv, d_\spu, n_\env)=(5,5,3)$.}
    \label{fig:averaged_results}
\end{figure}

\subsection{Default results} 

Table~\ref{main_results} shows that no method is able to achieve a performance close to the Oracle on any of the proposed problems.
The only exception are IRMv1 and ANDMask on \texttt{example1}.
This illustrates that current causal learning algorithms are unable to capture invariances, even in low-dimensional linear problems. 
The results averaged over the different environments are plotted in Fig.~\ref{fig:averaged_results}.

\subsection{Varying the number of environments}
\begin{figure}
\centering
    \includegraphics[width=\linewidth]{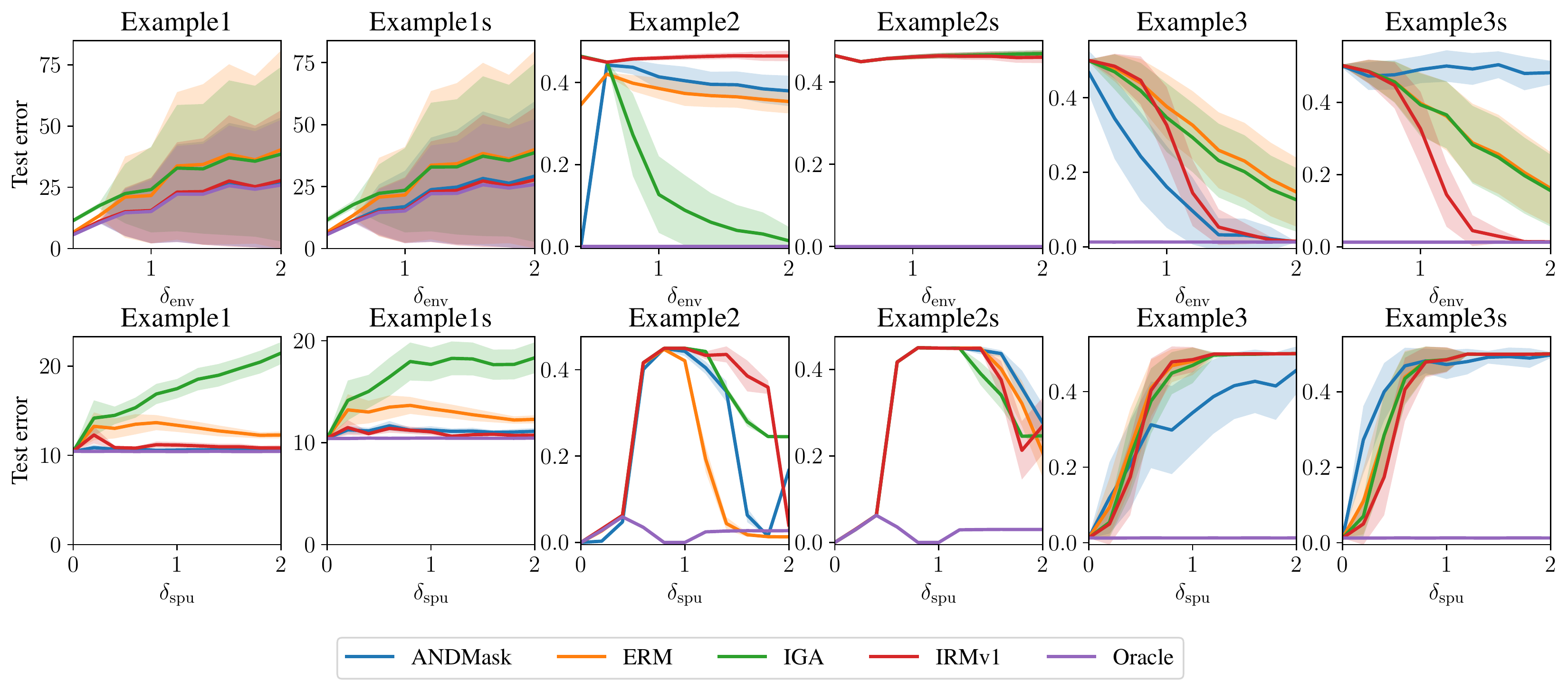}
    \caption{Test error averaged across environments for ANDMask, ERM, IGA, IRMv1 and Oracle on the unit-tests as (\textbf{top}) a function of the ratio $\delta_\env = {n_\env} / {d_\spu}$ at fixed dimensions $(d_\inv, d_\spu) = (5, 5)$; and as (\textbf{bottom}) a function of $\delta_\spu = {d_\spu} /{d_\inv}$ for $(d_\inv, n_\env) = (5, 3)$.
    }
    \label{fig:evolution_avg}
\end{figure}

What is the role of the number of environments $n_\env$ on generalization? We define the ratio $\delta_\env = \frac{n_\env}{d_\spu}$ between the number of environments and the number of spurious dimensions. We run experiments with the same procedure described above for $n_\env \in [2:10]$, and a fixed number of spurious dimensions $d_\spu = d_\inv = 5$. Figure~\ref{fig:evolution_avg} (\textbf{top}) show average test errors for all algorithms. Notably, IGA performs no better than ERM, except on \texttt{Example2} where it improves drastically when increasing $\delta_\env$; however, this increase bears no effect on the scrambled version \texttt{Example2s}.
    On \texttt{Example1} and \texttt{Example1s}, both ANDMask and IRMv1 approach closely Oracle's performances, while on \texttt{Example2} and \texttt{Example2s} simple ERM outperforms them. On the contrary, ANDMask and IRMv1 achieve good performances on \texttt{Example3}. They approach optimality for $n_\env \simeq d_\spu + 1$, since in this case no invariant boundary can solve the problem using $x_\text{spu}$ alone. IRMv1 performances do not suffer due to scrambling, while ANDMask collapses on \texttt{Example3s}.

\subsection{Varying the number of spurious dimensions}
We perform another ablation by fixing the number of environments $n_\env = 3$ and the number of invariant dimensions $d_\inv=5$, while varying the number of spurious dimensions $\delta_\spu = \frac{d_\spu}{d_\inv}$. We observe that for \texttt{Example1} and \texttt{Example1s}, ANDMask and IRMv1 do not suffer when adding spurious dimensions, while IGA crumbles as soon as a single spurious feature is added. As expected, on \texttt{Example3(s)} and \texttt{Example3s}, increasing the number of spurious dimensions while keeping the number of environments fixed decreases the performance of all algorithms. 
\texttt{Example2} and \texttt{Example2s} show the same phenomena for $\delta_\spu \leq 1$. 

\section{Outlook}
We propose a battery of ``unit-tests'' to surgically evaluate different types of out-of-distribution generalization abilities of machine learning algorithms.
While admittedly synthetic, our collection of problems attempts to cover a wide range of challenging distributional discrepancies that may arise across training and testing conditions.
We invite researchers to use and extend this set of problems to learn about the strengths and shortcomings of new algorithms in a transparent and standardized manner.

\clearpage
\newpage
\bibliographystyle{plainnat}
\bibliography{datasets}

\begin{thebibliography}{15}
\providecommand{\natexlab}[1]{#1}
\providecommand{\url}[1]{\texttt{#1}}
\expandafter\ifx\csname urlstyle\endcsname\relax
  \providecommand{\doi}[1]{doi: #1}\else
  \providecommand{\doi}{doi: \begingroup \urlstyle{rm}\Url}\fi

\bibitem[Alcorn et~al.(2019)Alcorn, Li, Gong, Wang, Mai, Ku, and
  Nguyen]{alcorn2019strike}
Michael~A Alcorn, Qi~Li, Zhitao Gong, Chengfei Wang, Long Mai, Wei-Shinn Ku,
  and Anh Nguyen.
\newblock Strike (with) a pose: Neural networks are easily fooled by strange
  poses of familiar objects.
\newblock \emph{CVPR}, 2019.

\bibitem[Arjovsky et~al.(2019)Arjovsky, Bottou, Gulrajani, and
  Lopez-Paz]{arjovsky2019invariant}
Martin Arjovsky, L{\'e}on Bottou, Ishaan Gulrajani, and David Lopez-Paz.
\newblock Invariant risk minimization.
\newblock \emph{arXiv preprint arXiv:1907.02893}, 2019.

\bibitem[Beery et~al.(2018)Beery, Van~Horn, and Perona]{beery2018recognition}
Sara Beery, Grant Van~Horn, and Pietro Perona.
\newblock Recognition in terra incognita.
\newblock \emph{ECCV}, 2018.

\bibitem[Ganin et~al.(2016)Ganin, Ustinova, Ajakan, Germain, Larochelle,
  Laviolette, Marchand, and Lempitsky]{ganin2016domain}
Yaroslav Ganin, Evgeniya Ustinova, Hana Ajakan, Pascal Germain, Hugo
  Larochelle, Fran{\c{c}}ois Laviolette, Mario Marchand, and Victor Lempitsky.
\newblock Domain-adversarial training of neural networks.
\newblock \emph{JMLR}, 2016.

\bibitem[Geirhos et~al.(2020)Geirhos, Jacobsen, Michaelis, Zemel, Brendel,
  Bethge, and Wichmann]{geirhos2020shortcut}
Robert Geirhos, J{\"o}rn-Henrik Jacobsen, Claudio Michaelis, Richard Zemel,
  Wieland Brendel, Matthias Bethge, and Felix~A Wichmann.
\newblock Shortcut learning in deep neural networks.
\newblock \emph{arXiv preprint arXiv:2004.07780}, 2020.

\bibitem[Kingma and Ba(2015)]{kingma2014adam}
Diederik~P Kingma and Jimmy Ba.
\newblock Adam: A method for stochastic optimization.
\newblock \emph{ICLR}, 2015.

\bibitem[Koyama and Yamaguchi(2020)]{koyama2020out}
Masanori Koyama and Shoichiro Yamaguchi.
\newblock Out-of-distribution generalization with maximal invariant predictor.
\newblock \emph{arXiv preprint arXiv:2008.01883}, 2020.

\bibitem[Li et~al.(2018)Li, Tian, Gong, Liu, Liu, Zhang, and Tao]{li2018deep}
Ya~Li, Xinmei Tian, Mingming Gong, Yajing Liu, Tongliang Liu, Kun Zhang, and
  Dacheng Tao.
\newblock Deep domain generalization via conditional invariant adversarial
  networks.
\newblock \emph{ECCV}, 2018.

\bibitem[Parascandolo et~al.(2020)Parascandolo, Neitz, Orvieto, Gresele, and
  Sch{\"o}lkopf]{parascandolo2020learning}
Giambattista Parascandolo, Alexander Neitz, Antonio Orvieto, Luigi Gresele, and
  Bernhard Sch{\"o}lkopf.
\newblock Learning explanations that are hard to vary.
\newblock \emph{arXiv preprint arXiv:2009.00329}, 2020.

\bibitem[Pearl(2009)]{pearl2009causality}
Judea Pearl.
\newblock \emph{Causality}.
\newblock Cambridge university press, 2009.

\bibitem[Peters et~al.(2015)Peters, B{\"u}hlmann, and
  Meinshausen]{peters2015causal}
Jonas Peters, Peter B{\"u}hlmann, and Nicolai Meinshausen.
\newblock Causal inference using invariant prediction: identification and
  confidence intervals.
\newblock \emph{arXiv preprint arXiv:1501.01332}, 2015.

\bibitem[Peters et~al.(2017)Peters, Janzing, and
  Sch{\"o}lkopf]{peters2017elements}
Jonas Peters, Dominik Janzing, and Bernhard Sch{\"o}lkopf.
\newblock \emph{Elements of causal inference}.
\newblock 2017.

\bibitem[Rosenfeld et~al.(2018)Rosenfeld, Zemel, and
  Tsotsos]{rosenfeld2018elephant}
Amir Rosenfeld, Richard Zemel, and John~K Tsotsos.
\newblock The elephant in the room.
\newblock \emph{arXiv preprint arXiv:1808.03305}, 2018.

\bibitem[Szegedy et~al.(2013)Szegedy, Zaremba, Sutskever, Bruna, Erhan,
  Goodfellow, and Fergus]{szegedy2013intriguing}
Christian Szegedy, Wojciech Zaremba, Ilya Sutskever, Joan Bruna, Dumitru Erhan,
  Ian Goodfellow, and Rob Fergus.
\newblock Intriguing properties of neural networks.
\newblock \emph{arXiv preprint arXiv:1312.6199}, 2013.

\bibitem[Vapnik(1998)]{vapnik1998statistical}
Vladimir Vapnik.
\newblock Statistical learning theory wiley.
\newblock \emph{New York}, 1998.

\end{thebibliography}

\end{document}